%% file: index.tex
\documentclass[journal]{IEEEtran}

\usepackage[style=ieee,maxcitenames=2,mincitenames=1,isbn=false,backend=biber]{biblatex}

\usepackage[utf8]{inputenc}
\usepackage[T1]{fontenc}
\usepackage[english]{babel}
\usepackage{csquotes}
\usepackage{textcomp}
\usepackage{gensymb}
\usepackage{mathtools}

\usepackage{amsmath,xparse}

\usepackage{array,tabularx,ragged2e,booktabs}

\usepackage{url}
\usepackage{graphicx}  
\usepackage{float}  
\usepackage{multirow}

\usepackage{color,soul} 

\usepackage{caption}
\usepackage{placeins}  
\usepackage{stfloats}

\AtEveryBibitem{%
  \clearname{translator}%
  \clearfield{url}%
  \clearfield{pagetotal}%
}

\begin{document}

\title{P-CRITICAL: A Reservoir Autoregulation Plasticity Rule for Neuromorphic Hardware}
        \author{
            \IEEEauthorblockN{
                Ismael Balafrej\IEEEauthorrefmark{1},
                Jean Rouat\IEEEauthorrefmark{1}
            }
            \IEEEauthorblockA{ \\
                NECOTIS Research Lab,
                Electrical and Computer Engineering Dep.,
                Université de Sherbrooke, Quebec, Canada\\
                \IEEEauthorrefmark{1}\{ismael.balafrej, jean.rouat\}@usherbrooke.ca}}

\markboth{}
{Shell \MakeLowercase{\textit{et al.}}: Bare Demo of IEEEtran.cls for IEEE Journals}

\maketitle

\begin{abstract}
    Backpropagation algorithms on recurrent artificial neural networks require an unfolding of accumulated states over time. These states must be kept in memory for an undefined period of time which is task-dependent. This paper uses the reservoir computing paradigm where an untrained recurrent neural network layer is used as a preprocessor stage to learn temporal and limited data. These so-called reservoirs require either extensive fine-tuning or neuroplasticity with unsupervised learning rules. We propose a new local plasticity rule named P-CRITICAL designed for automatic reservoir tuning that translates well to Intel's Loihi research chip, a recent neuromorphic processor. We compare our approach on well-known datasets from the machine learning community while using a spiking neuronal architecture. We observe an improved performance on tasks coming from various modalities without the need to tune parameters. Such algorithms could be a key to end-to-end energy-efficient neuromorphic-based machine learning on edge devices.
\end{abstract}


\section{Introduction}
\input{introduction.tex}

\section{Proposed approach}
\input{proposition.tex}

\section{Experiments and Results}
\input{experiments_and_results.tex}

\FloatBarrier


\section{Conclusion}
\input{conclusion.tex}

\FloatBarrier

\appendices

\section{Parameters}
\input{annex_parameters.tex}

\section*{Acknowledgment}

The authors would like to thank Intel for giving us access to the Loihi chip, for travel grants and guidance, along with Hydro-Québec, Université de Sherbrooke, Compute Canada and the Natural Sciences and Engineering Research Council of Canada (NSERC) for funding this research. We'd also like to thank the many reviewers and the developers of all the mentioned python libraries. Finally, we thank Simon Brodeur for his valuable insights on the CRITICAL plasticity rule.

\FloatBarrier

\printbibliography

\end{document}

%% file: introduction.tex
\IEEEPARstart{R}{eservoir} computing (RC), brought by both \citeauthor{Jaeger2001}~\cite{Jaeger2001} and \citeauthor{Maass2002}~\cite{Maass2002} respectively as the Echo State Network (ESN) and the Liquid State Machine (LSM), have been successful on various tasks. These models share a similar architecture and can be seen as a three-layer artificial neural network. The middle layer is connected with itself and is called reservoir. Only the last layer is trained using supervised methods, making this architecture efficient in training time and therefore interesting for the new emergent domain of neuromorphic computing (NC). Indeed, the recurrent connectivity of reservoirs can scale well with the parallelization brought by NC, while the fast training time can enable on-chip energy-efficient learning. Reservoirs are therefore suitable for memory dependent tasks without the need of expensive backpropagation-based training.


The core idea is that reservoirs should be sufficiently large to facilitate random temporal pairing of the incoming input signals, whilst the non-linearities and the recursiveness of the neurons should make any task linearly separable. This theoretical background is shared with concepts from hyperdimensional computing and vector symbolic architectures~\cite{Kleyko2017,Kanerva2009}. Once the data has been through the reservoir, the output layer should be able to uniquely identify the signals. While this architecture is promising, it comes at a hidden cost: the reservoir's connectivity, thereafter referred to as topology, alongside its weights must be chosen carefully during initialization. Failure to do so results in poorly performing reservoir networks. The topology choice and the various parameters of the network constitute a search space and therefore, optimization of these parameters is possible.

In order to navigate this search space to optimize the results, one must first find an adequate mesure of performance. There exist three types of criteria that can distinguish between a reservoir able to perform or not. The first and most straightforward is simply post training accuracy. The second type of criterion consists of \textit{a posteriori} methods~\cite{Legenstein2007,Oztuik2007,Verstraeten2009,Gorad2019} where all reservoirs in the search space are compared with a scoring method that requires simulation time, but not full-scale training. Lastly, \textit{a priori} criteria encompass methods that can create reservoirs without needing simulations~\cite{Yildiz2012,Hajnal2006}. The reservoir's parameters can be searched through an algorithm~\cite{Roeschies2009,Ferreira2009,Ju2013,Reynolds2019,Tian2020} and compared with one of these criteria.

These criteria range from high to low computational cost. More resources spent on optimization typically yield better performances. Yet, this idea conflicts with the premise of having fast trainable networks. We therefore propose P-CRITICAL, a plasticity rule for neuromorphic applications that partially remove the need for prior tuning of reservoirs. This simple plasticity rule is based on the concept of criticality, as used in~\cite{Kello2010,Brodeur2012,Stepp2015}, with a focus on neuromorphic hardware. We benchmarked P-CRITICAL with well-known machine learning classification tasks and observed increased performances in the context of liquid state machines.

The key contribution of this paper is P-CRITICAL, a new plasticity learning rule that can tune a reservoir to a stable and favourable regime while following the constraints of a modern neuromorphic chip, namely Intel's Loihi research test chip~\cite{Davies2018}. These constraints are set mostly by the inability of adding memory in the neurocores, which we circumvent by using more neurons. We present results coming from a CPU/GPU implementation using PyTorch~\cite{NIPS2019_9015} and with Loihi. We also propose a new optimization scheme for the hyperparameter space of small-world topology algorithms inspired from the eigenvalues spectrum of connectomes. This technique offers a valuable performance boost at a low computational cost while remaining task independent.

\section{Related Works}
The ESN and the LSM are mostly distinguished by their respective neuron models. ESNs use perceptron-like neurons with non-linear activation functions such as $sigmoid$ or $tanh$, while LSMs use biologically inspired neurons, often leaky-integrate-and-fire (LIF) neurons. In both cases, these architectures comprise three layers: an input layer $W_{I}$, a reservoir layer $W_{R}$ and an output layer $W_{O}$. Both $W_{I}$ and $W_{R}$ are not trained with supervised methods such as backpropagation. Many LSM users implement a non-spiking readout layer $W_{O}$ using machine learning methods~\cite{Soures2017,Reynolds2019,Moinnereau2018,Luo2018,Soures2019} or even $n$-layers formal neural networks~\cite{Tieck2018}. In this paper, we focus on the reservoir component and we use a single formal layer with a softmax activation function as the output layer.

\subsection{Optimization of the Reservoir}

While $W_{R}$ is not trained, many authors include unsupervised neuroplasticity rules~\cite{Liu2019,Xue2017,Jin2016,Luo2018,Moinnereau2018} as a way to either keep biological realism or to provide higher computational performances. Similarly, several studies looked at the initialization of $W_{R}$ in combination with various topologies such as small-world~\cite{Xue2017,Kawai2018,Kawai2019,Manevitz2010} and scale-free networks~\cite{Deng2007,Manevitz2010}. Furthermore, by using an orthogonal matrix for $W_{R}$, \citeauthor{Hajnal2006}~\cite{Hajnal2006} have shown an increased probability of finding a valid reservoir configuration for the task at hand in the context of ESNs.

In any case, one must follow a strict condition when selecting $W_{R}$ to achieve satisfactory results. Reservoirs are typically prone to two major problems as $W_{R}$ is recurrent: the explosion or the fading of the internal states during recursion. In both cases, the information will be lost, albeit the explosion problem is worst as it can create uncontrollable noise (similar to chaotic behaviour~\cite{Bertschinger2004}). That explosion can be solved for the ESN if $W_{R}$ is diagonally Schur stable; this is known as the echo state property (ESP)~\cite{Yildiz2012}. As explained in \citeauthor{Yildiz2012}~\cite{Yildiz2012}, a simple recipe for generating $W_{R}$ that satisfies the ESP is to create a positive random matrix $W$, scale it down using the spectral radius $\rho$ of $W$, and add inhibitory connections by changing the sign of any desired weights $w^{ij}$, considering $W = (w_{ij})$. This is, of course, equivalent to creating a random matrix and scaling it with $\rho(|w_{ij}|)$:

\begin{equation}
    W_{R} \coloneqq \frac{W_{R}}{\rho(|w^{R}_{ij}|)}
    \label{eq:echo_state_property}
\end{equation}

Although equivalents of eq. \ref{eq:echo_state_property} are widely used in the ESN literature, these methods do not translate to LSMs~\cite{Tieck2018,Verstraeten2007} and have proven to be insufficient as the sole \textit{a priori} criterion for reservoir performances~\cite{Alexandre2009,Oztuik2007,Verstraeten2007}. Several other metrics have been explored for quantifying performances based on \textit{a posteriori} dynamic analysis such as the Lyapunov exponent $\mu$~\cite{Legenstein2007}, the average state entropy~\cite{Oztuik2007}, the dynamic profile of the Jacobian of $W_{R}$~\cite{Verstraeten2009} or the approximate state space model~\cite{Gorad2019}. These methods allow for a more guided search on a reservoir's parameters. The most common approach is to use an evolutionary based search algorithm~\cite{Roeschies2009,Ferreira2009,Ju2013,Reynolds2019}. Similarly, \citeauthor{Tian2020}~\cite{Tian2020} showed improvements with a neural architecture search designed for LSM. Unfortunately, it remains impractical to iterate over some search space when NC applications are targeted unless one can find a fixed set of parameters that is task independent. Such a network seems unlikely when considering that the scaling of the weights must somewhat match the amplitude of the input for adequate memory fading. More importantly, in the context of LSMs, if $W_{R}$ is tuned to account for some high frequency spiking input, the reservoir will not respond correctly for sparser input activity and the information will die out quickly. For the ESN, one typically normalize the input to upper-bound the added activity in the reservoir, but spike trains with binary spikes have a fixed amplitude and normalizing the frequency isn't possible for real-time systems without prior knowledge of the task.

There are few proposed models that can tune what is called the branching factor $\sigma$ of $W_{R}$. For $n_{pre}$, the number of spikes for a neuron and $n_{post}$, the number of post-synaptic spikes for said neuron:

\begin{equation}
    \sigma = \frac{n_{pre}}{n_{post}}
\end{equation}

We define $\bar{\sigma}$ as the mean branching factor of the neurons in $W_{R}$. There are three defined regimes for $\bar{\sigma}$ known as subcritical when $\bar{\sigma} < 1$, critical when $\bar{\sigma} = 1$ and supercritical when $\bar{\sigma} > 1$~\cite{Beggs2007}. For the latter, the aforementioned problem of unconstrained activity rises up. These regimes are similar to what can be expressed by the ESP~\cite{Yildiz2012} or the Lyapunov exponent~\cite{Legenstein2007}. Amongst the motivation, a $\bar{\sigma}$ slightly below 1.0 can offer sufficient fading memory properties for RC while being close to reproducing \textit{in vivo} spike avalanches~\cite{Priesemann2014}. As such, a few models have been proposed for locally tuning $\sigma$ in a reservoir of spiking neurons~\cite{Kello2010,Brodeur2012,Stepp2015}. \citeauthor{Kello2010}~\cite{Kello2010}'s algorithm is memory-less in the sense that the branching factor is only considered in consecutive time steps (at $t$ and $t+1$). \citeauthor{Stepp2015}~\cite{Stepp2015} on the other hand tunes biologically inspired STDP-like plasticity rules to exhibit criticality behaviours. While it is true that branching factor tuning algorithms may increase computational power by bringing a network to the edge-of-chaos~\cite{Beggs2007,Brodeur2012,Kello2010,Stepp2015}, we focus on the assumption that reservoir dynamics must be adapted regardless to $\bar{\sigma}\approx 1-\epsilon$ - or slightly subcritical - in order to maintain readable states with adequate memory decay.

To sum up, it remains computationally expensive to adjust $W_{R}$ for edge computing applications. \textit{A priori} methods are limited and reservoirs often require task-specific optimizations; some type of adaptation is therefore necessary. As illustrated in this work, branching-factor-based algorithms could be the answer to that problem.

\subsection{Eigenvalues Spectrum}
While the weights of a reservoir must adapt themselves to the incoming data in LSMs, the reservoir's connectivity - or topology - can be fixed. This is because the weights of the network should already adapt themselves to the task, regardless of the topology. There still exist topologies better than others - take the simple case of a non-connected network: it won't perform well. A fully connected reservoir could be a valid solution since synaptic plasticity can decrease the weights down to zero, making unwanted connections obsolete. Alas, connections come at a computational cost. This is particularly true for neuromorphic devices or simulators optimized for sparse representation. Therefore, a suitable topology with minimal connectivity, yet enough connections to be able to perform over a range of tasks, is necessary.

Many authors empirically verified that biologically inspired topologies perform better than their completely random counterparts. \citeauthor{Manevitz2010}~\cite{Manevitz2010} showed that scale-free networks are more robust to noisy neuron models in the context of RC. Similarly, \cite{Xue2017,Kawai2018,Kawai2019,Deng2007,Davey2006} presented improvements on various tasks with either small-world or scale-free topologies. \citeauthor{Wijesinghe2019a}~\cite{Wijesinghe2019a} introduced the concept of liquid ensembles, which can be thought as a small-world topology with disjoint inner networks, that allowed them to speedup computation while still observing the increased performances of small-worlds.

Recent studies looked at the patterns in the eigenvalues spectrum, or rather in the probabilistic distribution of the eigenvalues of the normalized Laplacian of unweighted and undirected connectomes~\cite{DeLange2016,Wang2017}. This distribution seems to be consistent intraspecies, and bifurcations from said distribution are linked to improperly developed brains \cite{Wang2017}. Years of evolution seem to have led to a fairly consistent topology.

%% file: proposition.tex
Our approach is defined in two parts. First, we present P-CRITICAL, the synaptic plasticity rule in section \ref{section:proposition-pcritical}. We then present a topology optimization scheme in section \ref{section:proposition-topology}.

\subsection{Branching-Factor Adaptation of Reservoirs With Weight Updates}
\label{section:proposition-pcritical}

\begin{figure}[h]
    \begin{center}
        \includegraphics[width=6cm]{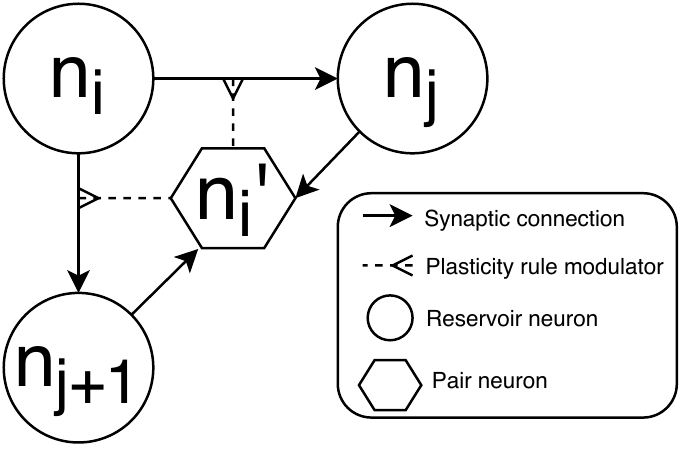}
        \caption{In this new adaptation rule, every neuron $n_i$ connected to neurons $n_j$ and $n_{j+1}$ of a reservoir is associated with its paired neuron $n_i'$. $n_i'$ approximates the branching factor of $n_i$ and generates spikes to either decrease or increase the weights of all outgoing connections of $n_i$.}
        \label{fig:pcritical}
    \end{center}
\end{figure}

\citeauthor{Brodeur2012}~\cite{Brodeur2012} introduced a neuron ensemble with a plasticity rule called CRITICAL that adapts the synaptic weights in a time-dependent manner. The locality of this plasticity rule incorporates new recorded states from the pre and post-synaptic neurons, referred to as pre and post-synaptic contributions. Unfortunately, such complex algorithmic behaviour for a plasticity rule is hard to reproduce in generic programmable neuromorphic chips such as Loihi. This is because the neuron model is embedded in the circuitry and therefore, adding new states in the model is not possible. Yet, we were able to translate the intended behaviour by recreating the adaptation part using what we refer to as paired neurons or $n_i'$. where $i$ is the index of a neuron in the reservoir. Each paired neuron is associated with a neuron of the reservoir and integrates the post-synaptic activity of neuron $i$. When the branching factor of neuron $n_i$ is above the targeted branching factor, $n'_i$ fires, causing a depreciation of all synaptic connections with $n_i$ as the presynaptic neuron. An overview of this concept can be visualized in figure~\ref{fig:pcritical} and equation~\ref{eq:pcritical}. We name this model P-CRITICAL (or Paired neuron CRITICAL).

We simulated that concept using the PyTorch~\cite{NIPS2019_9015} python library. We introduce two constants, $\alpha$ and $\beta$, that are used as learning rates for this new plasticity rule. The reservoir is initialized with a $W_R$ where the connectivity is decided with a distanced-based small-world topology (similar to~\cite{Brodeur2012}). We then connect the reservoir to the paired reservoir using weight tensor $W_R^T$ - the transpose of $W_R$. By doing so, every post-synaptic neuron of $N_i$ is connected to $N'_i$. When any paired neuron $N'_i$ spikes, the outgoing connections of $N_i$ are reduced by a factor $\alpha$ if initially positive and non-zero. At every time step, all non-zero positive weights are increased by $\beta$. The paired neurons $N'_i$ are simulated with the same LIF model as the reservoir neurons $N_i$, except for the voltage threshold $v_{th}$. The thresholds of paired neurons are increased by a factor $\alpha$ to target an average branching factor of one. Note that the threshold can be decreased or increased further to target a different branching factor for the reservoir. Finally, the amplitude of the weights are clipped between zero and one after the plasticity $W_R = \text{sign}(W_R) \circ \text{clip}(|W_R|, 0, 1)$ with $\circ$ as the Hadamard product. Unless stated otherwise, the typical $v_{th}$ chosen is one. A decaying exponential is added to enforce causality between pre and post synaptic spikes and weight changes, similar to short-term facilitation rules with constant $\tau$ in hebbian-like plasticity.
\begin{equation}
    \frac{dw_{ij}}{dt}=\beta-\alpha\delta(t-t_{n^{'}_i})e^\frac{-\Delta t_{ij}}{\tau}
    \label{eq:pcritical}
\end{equation}
$\delta$ represents the Dirac delta function centered at $t_{n^{'}_i}$, the paired neuron spike time. $\delta(t=0) = 1$ otherwise $\delta(t\neq 0) = 0$. $\Delta t_{ij}$ is the absolute time difference between the spike of neuron $n_i$ and $n_j$. In other terms, a connection is depreciated by $\alpha e^\frac{-\Delta t_{ij}}{\tau}$ whenever the pair neuron spikes. It is important to note that P-CRITICAL is unidirectional, meaning that all neuron weights could converge to zero if the pre-synaptic neuron is overly excited or if the learning rule overshoot its depreciation, which is why the weights of the reservoir are constantly growing by a fixed quantity $\beta$. The exponential can be approximated using a numerically decaying synaptic trace.

For all reservoirs, 20\% of the neurons are randomly chosen to be inhibitory. The weights are sampled from a uniform random distribution with range [0.2, 0.5[ for synapses coming from excitatory neurons and [0.1, 0.3[ for synapses coming from inhibitory neurons. All synaptic weights with an inhibitory pre-synaptic neuron are set to negative. The P-CRITICAL learning rule only affects connections coming from excitatory neurons, as inhibitory neurons cannot cause post synaptic spikes.

By design, we can translate P-CRITICAL easily on Intel's Loihi using their on-chip local learning rules. All weights are scaled from PyTorch's 23 bits mantissas (32 bits floating point implementation) to Loihi's 8 bits (+ 1 sign bit). As such, we converted the $\pm [0, 1[$ possible weight range to $\pm [0, 256[$. The only remaining constraint is that $\alpha$ and $\beta$ must be chosen to minimally affect the least significant bit of all weights while maintaining $\alpha > \beta$. This constraint adds a slight noise in the convergence of the weights for Loihi. The significance of this noise is discussed further in section \ref{section:results_validity}.

\subsection{Eigenvalues Spectrum Inspired Topology}
\label{section:proposition-topology}
Topology is another important aspect of a reservoir. It is known and well researched that some topologies are better than others for various tasks~\cite{Manevitz2010,Wijesinghe2019a,Xue2017,Kawai2018,Kawai2019,Deng2007,Davey2006}. But even within a topology choice, some hyperparameters are to be searched. We therefore propose a simple method of choosing an adequate set of parameters that is biologically inspired and task independent. We define our topology as small-world with a distance-based connectivity. We first create a three-dimensional Cartesian mesh grid of $\{x, y, z\}$ positioned neurons, equally separated with vectors of magnitude $s$ to their neighbours. We add a constant distance $p$ between every group of $j$ neuron in all orthogonal directions. This results in $\lfloor\frac{n}{j}\rfloor^3$ mini-reservoirs of size $k^3$ neurons assuming $W_R \in \!R^{n,n}$ and $n \equiv 0 \mod j$. The constant $j$ can either be represented as a numerical constant or a vector $\mathbf{j}\in\!R^3$ if the number of neurons in each axis of a mini-reservoir is not the same; resulting in non-cubic mini-reservoirs.

The adjacency matrix can be generated by randomly connecting neurons based on their Euclidean distance $D$, as done in~\cite{Maass2002}. The probability $P$ of connection between neuron $a$ and neuron $b$ is given by:
$$
    P = C \cdot e^{-\frac{D(a, b)}{\lambda}}
$$
Where $C$ is the maximum connection probability and $\lambda$ is a control parameter which we refer to as a Euclidean distance divisor.

By looking at the eigenvalues spectrum of the macaque as presented in~\citeauthor{DeLange2016}~\cite{DeLange2016}, we manually tuned $s$, $p$, $C$ and $\lambda$ to minimize the Kullback–Leibler divergence with the simulated topology's eigen spectrum. We obtained a good approximation with values $s=40$, $p=1460$, $C=0.11$ and $\lambda=635$. From a topological perspective, our reservoir will yield a more similar macroscopic structure to what is seen in the brain. Therefore, this method uses million of years of evolution in connectomes to enhance our reservoir's topology.

The input matrix $W_I$ is simply a permutation matrix\footnote{A permutation matrix can be created by randomly permuting the rows of the identity matrix. This is equivalent to connecting each input neuron to a unique reservoir neuron (one-for-one).} multiplied by a constant weight $w^I_{ij} \gg v_{th}$. By doing so, we remove any need to consider the input weights distribution. As the weights are much larger than the threshold voltage, any input spike will create one reservoir spike. We therefore consider the input neurons to be within the reservoir and plastic connections can act immediately. $1$ to $n$ connections can be created by changing $n-1$ zeros into ones in each row of $W_I$ before the permutation operation.

For classification tasks, we bin the reservoir spikes into $R_{\text{output}}$ by counting the spikes in fixed lenghts of time and train a weight matrix $W_O$. $W_O$ is trained with backpropagation using PyTorch's cross entropy loss function. We use a batch normalization layer $\text{b}_\text{n}$~\cite{Ioffe2015a} in between the reservoir output and the single layer classifier. Accuracies are calculated from labels $y$ with $\sum(y=\text{argmax}(\text{b}_\text{n}(R_{\text{output}})W_O))$.

%% file: experiments_and_results.tex
We tested our method against two well-known datasets of the machine learning and spiking neural network community: N-MNIST~\cite{Orchard2015} and N-TIDIGITS~\cite{Anumula2018}. Both of these datasets were created using event-based sensors from previously recorded data. N-MNIST comes from the saccadic presentation of the well-known handwritten digit recognition dataset MNIST to the event-based camera \textit{ATIS sensor}~\cite{Posch2011a}. N-TIDIGITS was recorded from TIDIGITS, an audio representation of spoken digits, using the spiking silicon cochlea sensor \textit{CochleaAMS1b}~\cite{Chan2007a,Anumula2018}. Successful training on these datasets could be a significant step to end to end training of low energy event-based hardware. As mentioned, the main goal of P-CRITICAL is to tune a reservoir to the input spike train representation as to offer stability. While many publications present optimized reservoir parameters for the task in hand, we demonstrate that P-CRITICAL can compensate for bad sets of initial parameters and to some extent a bad initialization - i.e. an initialization that is not suited for a specific task. All parameters for the various experiments are attached in appendix \ref{section:parameter}.

\subsection{Validity of the Model}
\label{section:results_validity}
\begin{figure}
    \begin {center}
        \includegraphics[width=0.48\textwidth]{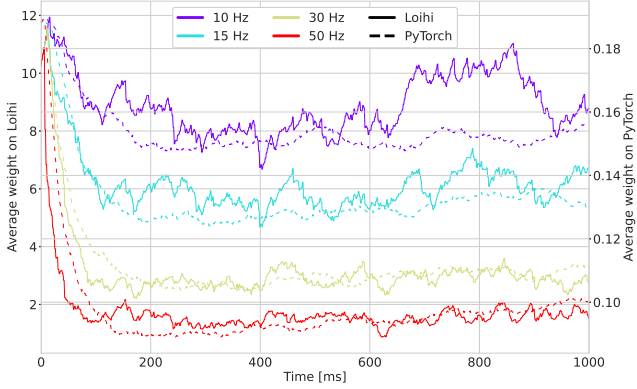}
        \caption{Autoregulation of the weights using the \mbox{P-CRITICAL} plasticity rule with random input and a target branching factor of 1. The average weight is shown as a function of time. There are 170 neurons in the 1s input spike train with a one-to-one connectivity to the 512 reservoir neurons. The input spike train was sampled from a Poisson distribution of labelled frequencies. It is observed that P-CRITICAL is regulating the weights in function of the spiking input frequency. Results are presented for both Loihi and PyTorch implementations.}
        \label{fig:weight_adaptation}
    \end {center}
\end{figure}

To test the behaviour of the P-CRITICAL model, we begin with a simple reservoir of $512$ neurons and $170$ neurons with Poisson spiking activity as input. The input neurons are connected in a one-to-one fashion to the reservoir. We vary the random input frequency from 10 to 50 Hertz. We aim for the reservoir to have a mean branching factor $\bar{\sigma} = 1$. The small-world topology constant $j=4$. The results are shown in figure \ref{fig:weight_adaptation}. As expected, the weights of the reservoir converge according to the input frequency to maintain a steady activity in the reservoir. For higher-frequency inputs, the average weight should be smaller while doing the opposite for smaller inputs. Similarly, we re-created the experiment on the Loihi chip and, as expected, we obtained similar results with the P-CRITICAL rule where we found that the final average weight is inversely proportional to the input spike frequency.

\begin{figure*}[bp]
    \begin{center}
        \includegraphics[width=.8\textwidth]{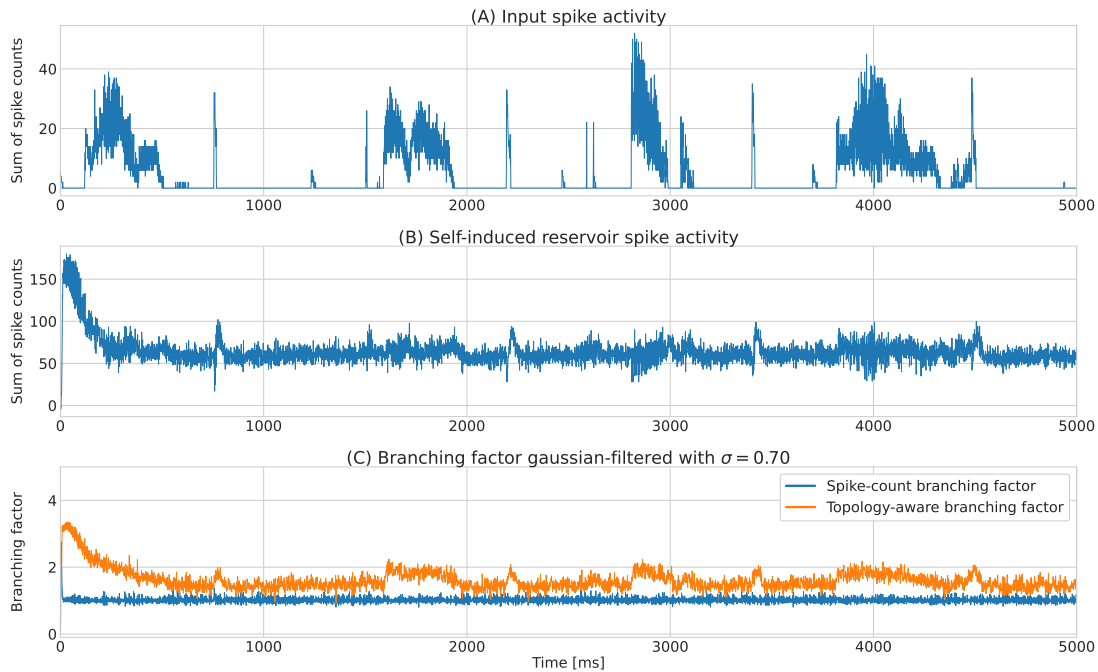}
        \caption{Spike activity for 5000 ms of randomly chosen samples from the N-TIDIGITS dataset. Plot A shows the input spikes mapped to the 512 neurons inside the reservoir and summed. Each of the 64 input features was connected to two reservoir neurons. Plot B shows the spike count summed of the reservoir in which the input spike raster was subtracted, therefore showing the self-induced activity inside the reservoir. Plot C shows the branching factor, estimated over the self-induced reservoir activity with two different methods as done in~\cite{Stepp2015}: the total spike count and a topology-aware method. This branching factor estimation is Gaussian filtered over time with constant $\sigma=0.7$. The total spike count method consists of dividing the total spike count at t+1 divided by the total spike count at t. The topology-aware method works similarly, but spike counts are computed locally and the resulting branching factors are averaged over all neurons in the reservoir. Both estimation methods are presented for excitatory neurons only.}
        \label{fig:ntidigits_spikes}
    \end{center}
\end{figure*}

We proceed with an evaluation of the branching factor similar to~\citeauthor{Stepp2015}~\cite{Stepp2015}. We first subtract the input spike train, mapped to the reservoir's dimension by $W_I$, from the reservoir's spike train. This way, we ensure that the branching factor computation methods only consider the self-induced activity within the reservoir. We compute a local branching factor estimation where every neuron's post-synaptic activity is summed and divided by its pre-synaptic activity in terms of spike count. We then average this value for every excitatory neurons in the reservoir. This topology-aware method will overshoot slightly the global branching factor as post-synaptic spikes can be counted multiple times by pre-synaptic neurons. We then estimate the global branching factor with the total number of spikes at $t+1$ divided by the total number of spikes at time $t$ for excitatory neurons. For these tests, we use 5 seconds of continuous activity randomly sampled from the N-TIDIGITS dataset. Both methods revealed a fairly consistent branching factor of 1 with the P-CRITICAL learning rule, after a small adaptation period. Spike activity and branching factor estimations are illustrated in figure \ref{fig:ntidigits_spikes}.

\begin{figure}
    \begin {center}
        \includegraphics[width=0.48\textwidth]{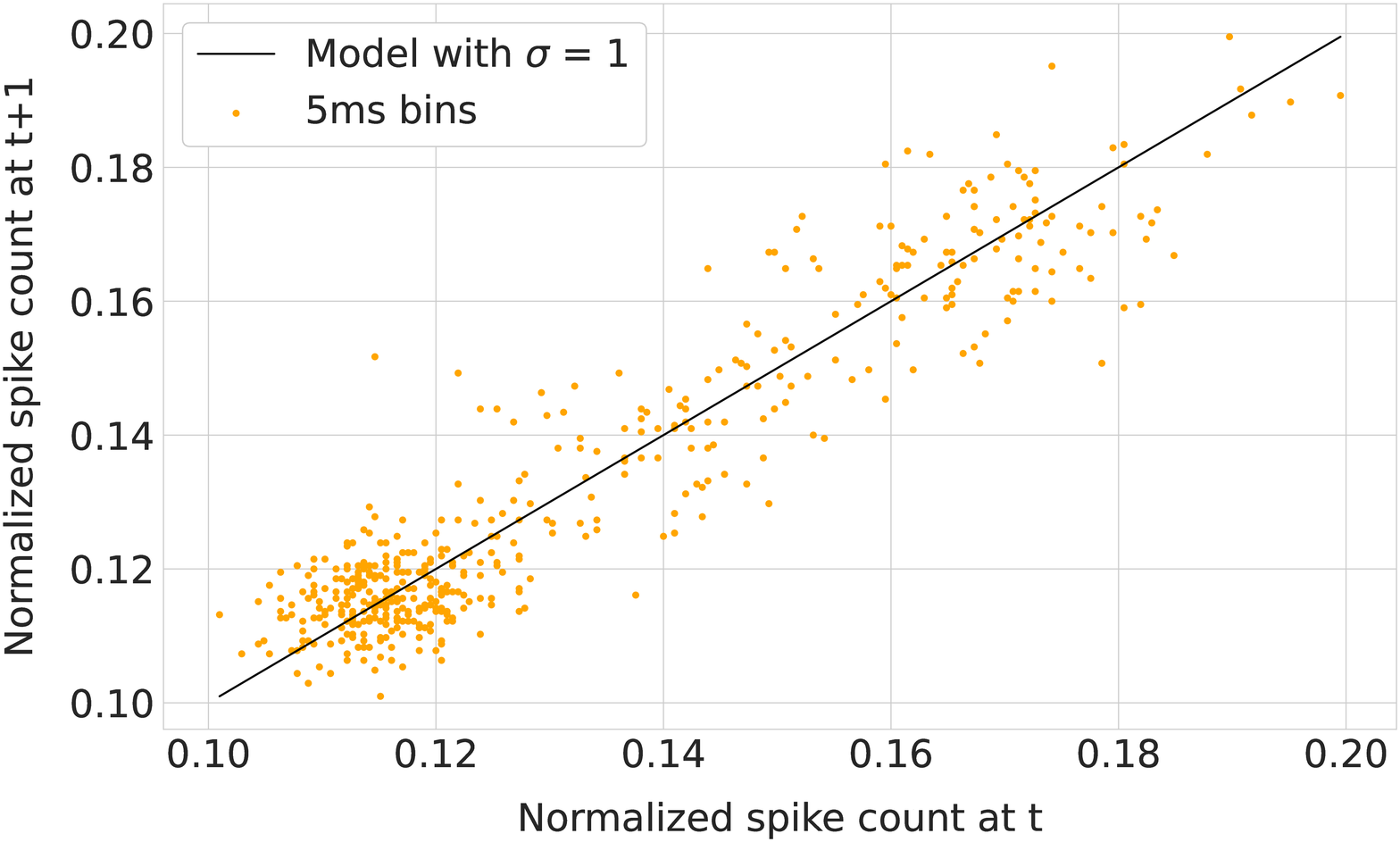}
        \caption{Poincaré visualization of the spike count time-binned with $T_{\text{bins}}=5$ milliseconds. A slope of one was added as a comparison for a model with $\bar{\sigma}=1$. The spikes come from the self-induced reservoir activity after presentation of the N-TIDIGITS dataset. We first tuned the reservoir over 2500 milliseconds of continuous input. We then simulate the model for another 2500 milliseconds, which resulted in the presented spike counts. We used a small-world reservoir of 512 neurons with constant $j=4$.}
        \label{fig:ntidigits_spikes_poincare}
    \end {center}
\end{figure}

Finally, we also compare the time-binned spike counts from the self-induced activity in a Poincaré plot in figure \ref{fig:ntidigits_spikes_poincare}. We once again used 5s of activity from the N-TIDIGITS datasets where features were randomly connected to two reservoir neurons each. We removed the first 2.5s of the spike train to be sure that the reservoir had converged to the target branching factor of one. We then compute the spike counts using 5 ms bins and plot these counts for consecutive time periods. We compare this with a model of slope one, which represents a $\bar{\sigma}=1$.
We observe that the P-CRITICAL enabled reservoir can adequately maintain a branching factor of one.

\subsection{Real-World Tasks}
We then compare randomly initialized reservoirs with P-CRITICAL on N-TIDIGITS and N-MNIST. All experiments are averaged over 5 executions using different random seeds and the standard deviation is presented. We also use identical LIF parameters for both experiments as we would expect in a generic reservoir-based NC chip, even though they come from different sensory representations. All parameters are reported in appendix \ref{section:parameter}.

\subsubsection{Speaker-Independent Audio Digit Classification}
\begin{figure}
    \begin {center}
        \includegraphics[width=0.48\textwidth]{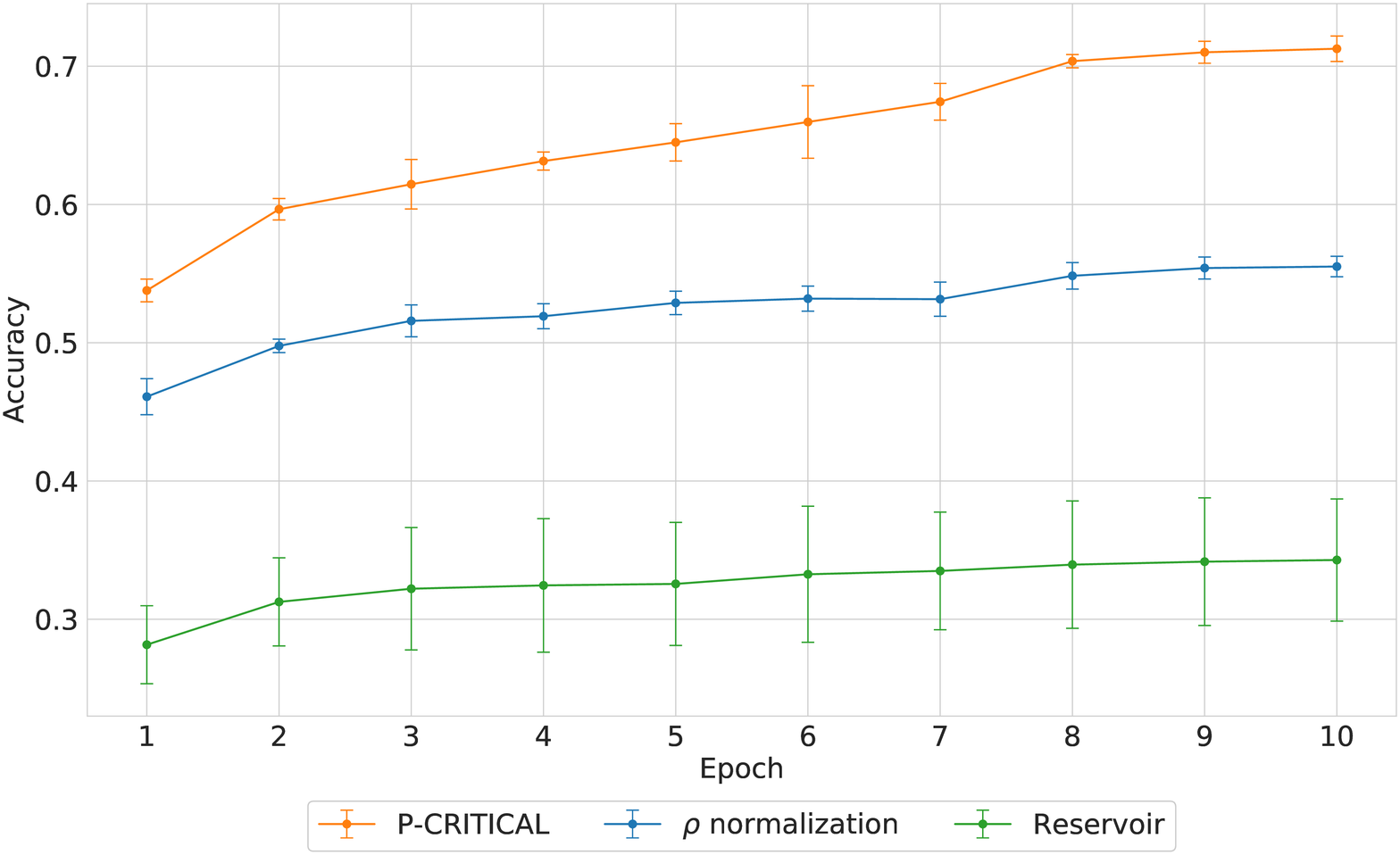}
        \caption{Test accuracy as a function of the number of epochs for the N-TIDIGITS classification task. No hyper-parameters were tuned for this task. Classification results are sampled 5 times for each method. All initialization constants are presented in appendix \ref{section:parameter}. The spectral radius $\rho$ normalization of the weights has an expected improved accuracy versus a completely random initialization. However, P-CRITICAL enabled reservoirs surpassed all other methods.}
        \label{fig:ntidigits_classification_results}
    \end {center}
\end{figure}

For the N-TIDIGITS classification task, we used a 512 neurons reservoir with small-world topology constant $j=4$. For comparison, we run the same sets of experiments with no plasticity and no tuning of the initial parameters and we also use the spectral radius $\rho$ normalization from eq. \ref{eq:echo_state_property}. We run our model for 10 epochs, and we use a batch size of 32 samples when training the output layer. Only the single digit samples of the dataset were used for training. We use the Adam~\cite{Loshchilov2019} optimizer with a learning of $10^{-3}$. As expected, reservoirs that were tuned using spectral radius normalization outperformed random reservoirs. We observe, however, an increased accuracy on the test set with all experiments where P-CRITICAL plasticity was enabled. We obtained with P-CRITICAL an average accuracy of $71.26\pm0.92\%$ (figure \ref{fig:ntidigits_classification_results}). We executed the exact same reservoir experiment on the Loihi research chip and observed a 64.1\% accuracy. As the reservoir is slightly more noisy because of bit-depth, we ran the model for 10 more epochs when using the loihi model with a weight decay of $10^-2$. Finally, optimizing the eigenvalues spectrum offered all tested reservoirs a mean accuracy boost of 16.77\% on N-TIDIGITS at no task-specific optimization cost. To the knowledge of the authors, this is the first LSM-based reported accuracy for the N-TIDIGITS dataset.

\subsubsection{Handwritten Digit Classification}
\begin{table*}
    \centering
    \begin{tabularx}{\linewidth}{ccXc}
        \hline
        Model & Reservoir size & \multicolumn{1}{c}{Details} & Mean accuracy (\%) \\ [0.5ex]
        \hline\hline
        \citeauthor{Iranmehr2019}~\cite{Iranmehr2019} & 625 & Unoptimized reservoir & 91.48 \\
        \hline
        \citeauthor{Iranmehr2019}~\cite{Iranmehr2019} (GA) & 625 & Optimized reservoir & 92.56 \\
        \hline
        \citeauthor{Iranmehr2019}~\cite{Iranmehr2019} (GA+HFC) & 625 & Optimized reservoir with a 120 neurons hidden FC layer& \textbf{98.38} \\
        \hline
        \citeauthor{Guo2020}~\cite{Guo2020}\footnote{Only a 10k samples subset of N-MNIST was used (+10k in testing).} & 1000 & This work focus on input compression for smaller reservoirs & 91.67 \\
        \hline
        \citeauthor{Thiele2018}~\cite{Thiele2018} & - & This work uses an unsupervised STDP trained CNN & 95.77 \\
        \hline
        P-CRITICAL (this work) & 8640 & Unoptimized reservoir & \textbf{95.22} \\
        \hline
        P-CRITICAL (quadrant method) & 1156 & Unoptimized reservoir, on-chip & 88.61 \\
    \end{tabularx}
    \caption{Comparison of several models benchmarked on N-MNIST. All except~\cite{Thiele2018} are based on the LSM architecture. All the models presented use fast unsupervised training layers with a trained readout layer.}
    \label{table:nmnist_results}
\end{table*}

For N-MNIST, we use a 8640 neurons reservoir (with $\mathbf{j}=\begin{bmatrix}4&4&3 \end{bmatrix}$) with PyTorch. In all cases, only the ON polarity of the input spike trains as available in the N-MNIST dataset was kept. For the readout layer, we used the Adam optimizer with amsgrad~\cite{Reddi2018}, a learning rate of $1e-5$ and a batch size of 10. We observe a $95.22\pm0.09\%$ accuracy on the test data. As N-MNIST is more substantial in the amount of data, only 1 epoch through the whole dataset was necessary to achieve these results. We conducted a second faster experiment with only 1156 neurons in the reservoir. To do so, the 3D input spike train of shape 34x34xTime was split into sub-spike-trains, or quadrants, of shape 17x17xTime. We ran the experiment on the Loihi chip with P-CRITICAL and obtained an accuracy of 88.61\%. We refer to this second experiment as the quadrant method. We compare our method with other mostly unsupervised approaches on N-MNIST in table~\ref{table:nmnist_results}. P-CRITICAL was able to surpass other unoptimized reservoir-like methods. As mentioned, no hyperparameters optimization was done and the number of neurons was selected on the assumption that the size of the reservoir had to be larger than the input spike train.

\subsection{Neuromorphic efficiency}

\begin{table*}
    \centering
    \begin{tabularx}{\textwidth}{XXXX|XX}
    \hline
    & \multicolumn{3}{c}{Power (mW)} & \multirow{2}{*}{Time/timestep ($\mu$s)} & \multirow{2}{*}{Energy per timestep ($\mu$J)} \\

    & Static (Idle)   & Dynamic   & Total   &  &  \\

    \hline\hline

    Loihi neurocores  & 0.91 $\pm$ 0.10       & 18.3  $\pm$ 0.1       & 19.2 $\pm$ 0.2      & 17.52 & 0.336  $\pm$ 0.004                                                             \\
    \hline
    Intel i7-9750H                          & 5 380 $\pm$ 40       & 46 000  $\pm$ 2 000       & 51 000 $\pm$ 2 000 & 880 & 45000 $\pm$ 2000 \\
    \hline
    \end{tabularx}

    \caption{Comparison of energy consumption and speed of P-CRITICAL implementations for a 512 neurons reservoir. CPU efficiency was measured using Intel SoC Watch on Linux with kernel version 5.4.0-7634, Python 3.8.1 and PyTorch 1.4.0. Loihi efficiency was measured using Nx SDK version 0.9.5 on Nahuku 32 board ncl-ext-ghrd-01. Spikes were generated using input neurons with a bias current to simulate a 40 Hz input frequency on Loihi in order to avoid I/O latency. Intel's Loihi research chip has shown major improvements in both power and time efficiency for P-CRITICAL when compared to a conventional CPU.}
    \label{table:energy_consumption}
\end{table*}

A 512 neurons reservoir with P-CRITICAL only takes about 2 to 3 neurocores on Loihi depending on the connectivity, out of a possible 128 cores per chip. We benchmarked our reservoir running on 2 neurocores on Loihi with a chosen power-efficient CPU: an Intel i7-9750H. The CPU ran the PyTorch implementation. Such a network takes on average 0.88 ms per timestep to run on PyTorch. This model is therefore 1.13 times faster than our simulated timestep of 1 ms on PyTorch. In comparison, the same reservoir takes 17.52 $\mu$s on Loihi. In contrast to PyTorch, this is about 50 times faster. When scaled to all 128 cores of a single Loihi chip (64x 512 neurons reservoirs), this amount of time is only increased to 19.75 $\mu$s per timestep because of the parallel nature of the chip. We ran all efficiency experiments using Nx SDK version 0.9.5 on Nahuku 32 board ncl-ext-ghrd-01 with power probing. We also benchmarked power-efficiency for both implementations. The PyTorch version consumes 46W of dynamical power. In comparison, the Loihi implementation only takes 17.3 mW. This is more than three order of magnitudes more power efficient than with the CPU. Table \ref{table:energy_consumption} shows a breakdown of energy and time consumption with comparison to a chosen power-efficient CPU: Intel i7-9750H.

%% file: conclusion.tex
P-CRITICAL achieved its goal by tuning the branching factor of various reservoirs. The plasticity rule was able to offer a stable activity when connected to various raw input spike trains. As figure \ref{fig:ntidigits_spikes} demonstrated, even with a sparse input, the reservoir can maintain a fairly constant activity. By doing so, the reservoir will not suffer from sub or super criticality. Furthermore, this branching factor model should allow edge of chaos behaviour, maximizing the computing power and memory retention of the reservoir~\cite{Kello2010}. The plasticity rule was able to increase the test accuracy of unoptimized reservoirs for various high-level tasks coming from different sensory inputs that were captured with event-based sensors. We aim for P-CRITICAL to extend current reservoir computing methods such that they can be implemented on a neuromorphic processor and offer low-power edge devices the ability to train without requiring extensive computation or cloud server access.

Reservoir computing is a good alternative to RNNs for faster training times, and plasticity-enabled reservoirs are well suited for neuromorphic engineering applications as they are slow on conventional computer architecture. This is because plasticity rules often prevent batching of the input data while the recurrent dynamics of reservoirs forces sequential simulations.

This new model was compared on both a CPU following the von Neumann architecture and the Loihi neuromorphic research chip. Both in time and power efficiency, Loihi was able to outperform its counterpart by orders of magnitude.

Furthermore, we created a new topology optimization scheme dans is task independent and based on the eigenvalues spectrum of connectomes. This approach is a simple way of tuning the hyperperameters related to topology. Having a fixed topology would further help hardware implementations as it removes the need to support generic topologies.

In conclusion, we presented P-CRITICAL, a plasticity rule created for the autoregulation of reservoirs that tunes the branching factor to a target value. The plasticity rule was designed and adapted from recent literature~\cite{Brodeur2012} with Intel's Loihi as a target platform. With the hardware constraints in mind, we developed a plasticity rule able to successfully increase the computational power of reservoirs in liquid state machines. We believe that this will be a key component for end-to-end energy-efficient machine learning algorithms on edge devices. In future works, we hope to combine our reservoir-plasticity method with state-of-the-art LSM readout layers~\cite{Zhang2015} that can account for spike dynamics.

%% file: annex_parameters.tex
\label{section:parameter}

All simulations were executed with a numeric differential step size $dt = 1$ ms in both PyTorch and Loihi. Although similar in most cases, both PyTorch and Loihi values are presented.

\begin{center}
    \begin{table}
        \caption{Current-leaky-integrate-and-fire generic constants}
        \begin{tabularx}{\linewidth}{ccc*2{>{\RaggedRight\arraybackslash}X}}
        \hline
        Symbol & PyTorch & Loihi & Description \\ [0.5ex]
        \hline\hline
        $\tau_v$ & \multicolumn{2}{c}{30 ms} & Membrane potential decay constant \\
        \hline
        $\tau_i$ & \multicolumn{2}{c}{1 ms}  & Membrane current decay constant \\
        \hline
        $v_{\text{reset}}$ & \multicolumn{2}{c}{0} & Membrane reset voltage \\
        \hline
        $v_{\text{threshold}}$ & 1.0 & 256 & Membrane threshold voltage \\
        \hline
        $T_{\text{refractory}}$ & \multicolumn{2}{c}{2 ms} & Refractory period \\
        \hline
        \end{tabularx}
    \end{table}
\end{center}

\begin{center}
    \begin{table}
        \caption{Small-world topology constants}
        \begin{tabularx}{\linewidth}{ccc*2{>{\RaggedRight\arraybackslash}X}}
        \hline
        Symbol & PyTorch & Loihi & Description \\ [0.5ex]
        \hline\hline
        $s$ & \multicolumn{2}{c}{40} & Distance between neurons \\
        \hline
        $p$ & \multicolumn{2}{c}{1460} & Distance increment between small-worlds \\
        \hline
        $C$ & \multicolumn{2}{c}{0.11} & Maximum probability connection \\
        \hline
        $\lambda$ & \multicolumn{2}{c}{635} & Euclidean distance divisor constant \\
        \hline
        $W_R^{{\text{Excittory}}}\sim$ & [0.2, 0.5[ & [51.2, 128[ & Uniform distribution range of excitatory weights \\
        \hline
        $W_R^{{\text{Inhibitory}}}\sim$ & [0.1, 0.3[ & [25.6, 75.8[ & Uniform distribution range of inhibitory weights \\
        \hline
        \end{tabularx}
    \end{table}
\end{center}

\begin{center}
    \begin{table}
        \caption{P-CRITICAL constants}
        \begin{tabularx}{\linewidth}{ccc*2{>{\RaggedRight\arraybackslash}X}}
        \hline
        Symbol & PyTorch & Loihi & Description \\ [0.5ex]
        \hline\hline
        $\alpha$ & 1e-2 & 2 & Learning rate \\
        \hline
        $\beta$ & 1e-5 & 0.25 & Increment constant \\
        \hline
        $\tau_v'$ & \multicolumn{2}{c}{5 ms} & Membrane potential decay constant for paired neurons \\
        \hline
        $\tau_i'$ & \multicolumn{2}{c}{0 ms} & Membrane current decay constant for paired neurons \\
        \hline
        \end{tabularx}
    \end{table}
\end{center}

\begin{center}
    \begin{table}[h!]
        \caption{Time-binned read-out layer constants}
        \begin{tabularx}{\linewidth}{ccc*2{>{\RaggedRight\arraybackslash}X}}
        \hline
        Symbol & PyTorch & Loihi & Description \\ [0.5ex]
        \hline\hline
        $T_{bins}$ & \multicolumn{2}{c}{60 ms} & Size of the time bins \\
        \hline
        \end{tabularx}
    \end{table}
\end{center}